# Design and Implementation of a Virtual 3D Educational Environment to improve Deaf Education

Abdelaziz Lakhfif
LRSD, Computer Sciences Department, Faculty of Sciences
Ferhat Abbas Setif 1 University
Setif, Algeria
abdelaziz.lakhfif@univ-setif.dz

*Abstract*—Advances in NLP, knowledge representation and computer graphic technologies can provide us insights into the development of educational tool for Deaf people. Actual education materials and tools for deaf pupils present several problems, since textbooks are designed to support normal students in the classroom and most of them are not suitable for people with hearing disabilities. Virtual Reality (VR) technologies appear to be a good tool and a promising framework in the education of pupils with hearing disabilities. In this paper, we present a current research tasks surrounding the design and implementation of a virtual 3D educational environment based on X3D and H-Anim standards. The system generates and animates automatically Sign language sentence from a semantic representation that encode the whole meaning of the Arabic input text. Some aspects and issues in Sign language generation will be discussed, including the model of Sign representation that facilitate reuse and reduces the time of Sign generation, conversion of semantic components to sign features representation with regard to Sign language linguistics characteristics and how to generate realistic smooth gestural sequences using X3D content to performs transition between signs for natural-looking of animated avatar. Sign language sentences were evaluated by Algerian native Deaf people. The goal of the project is the development of a machine translation system from Arabic to Algerian Sign Language that can be used as educational tool for Deaf children in algerian primary schools.

I. INTRODUCTION

In the Sign language Generation (SLG) field, the marriage of NLP lexical semantic resources, knowledge representation and computer graphic technologies has taken the development of accessibility and assistive technologies (AT) an important step forward and has open new perspectives to make learning tools really intuitive and useful for the deaf community. In Algeria, special schools for deaf children, use images and textbooks to teach young pupils with hearing disabilities, whereas most of deaf children in primary schools have difficulties with reading and writing text and prefer visual methods.

Till now, educational materials for pupils with hearing disabilities do not reach the richness and the comprehensibility of usual educational supports of the same levels for normal pupils. One accurate and a low cost alternative solution to combat this deficiency is the use of computer technologies to translate usual educational courses into Sign language performed by virtual characters (avatars), taking full advantages of existing educational materials and enabling Deaf pupils to experience the same contents by providing them with full access in educational settings. Such a promising approach to virtual education enhances the acceptability by Deaf children and increases their comprehensibility capacity.

Three-Dimensional Virtual Reality Environments (3D-VRE) have become increasingly preeminent tool for special education systems and programs such as those dedicated for deaf and hearing-impaired people. The capability of generating, virtually and automatically, any model of real world objects which can be animated in interactive mode allows educational contents based on 3D-VRE to be more attractive to pupils with hearing disabilities. 3D-VRE provides a huge range of objects and resources (virtually instantiated) as well as generates parametric behaviors that can simulate real world scenes [30].

One of the most important advances in 3D-VRE has been the wide speared acceptance of the Web3D open standards. Web3D open standards aim at providing 3D-VRE based applications with interactivity and deployment and allowing for the principle of "writing once, publishing everywhere". Thus, in order to take advantages of both collaborative interactivity and contents deployment provided by Web technologies, linguistic features of Signs are represented and stored in H-Anim and X3D format, allowing standard web browsers to display signing sentences in consistent and ordinary way. VRML and its successor X3D standards are a powerful and general-purpose format that provides various enhanced built-in features for defining objects and animations.

Besides providing 3D Sign language dictionary, our attempt is intended to provide an educational tool that meets Deaf children's learning requirements and includes Sign Language sentences in Web3D standards and technologies, increasing hence accessibility, interactivity and collaboration among Deaf community. The use of Web3D standards to sign language synthesis allows unlimited X3D and H-Anim based avatars to perform the generating written text-to-sign language translation in same manner.

In the present paper, we present our attempts to design and implement virtual environment architecture, building on a range of state-of-the-art research techniques and resources, and that support Arabic text learning using Sign Language as a primary communication means. The tool is devoted for pupils with hearing disabilities in classrooms in primarily level education. This work aims to implement a gestural communication tool for the Algerian deaf pupils to facilitate their access to textual books written in standard Arabic.

The rest of this paper is organized as follows: (1) we present the state-of-the-art of virtual avatars in MT systems in Section 2. Theoretical backgrounds of this work will be discussed in Section 3. The architecture of the proposed systems is briefly described in Section 4. In Section 5, we present first results and conclusions.

## II. RELATED WORKS

The last decade has seen a great maturation in Sign Language processing in order to promote deaf peoples community. Many Sign Language machine translation systems are developed for several Sign languages in the world [1], [3], [5], [6], [7], [8], [9], [33], [34], [35], [36] and [10].

ViSiCAST project is an Avatar-based Sign Language system for translating English text into British Sign Language [1], [2]. The approach uses a CMU Link Parser to analyze an input English text, and then uses a Prolog declarative clause grammar rules to convert this linkage output into a Discourse Representation Structure (DRS). The ViSiCAST system uses a more detailed ASL animation script which includes phonological information encoded in the Signing Gesture Markup Language [4]. Following the success of VisiCAST project and other similar projects, continuing efforts in the field of virtual 3D environments based systems for education [17], [19], and [20], 3D-VRE have attracted considerable attention in learning tools development for pupils with hearing disabilities [15], [16], and [18], [24], thanks to its capabilities to generate synthetic, highly interactive 3D spatial pedagogical contents that represent real or non-real objects.

## III. BACKGROUND AND MOTIVATIONS

Sign Language is the primary means of communication and the appropriate language used by deaf people to communicate just as spoken languages for hearing people. Modern linguistic studies have proved that Sign Languages are full and complex languages with rich expressive capabilities [25], [26] and [27].

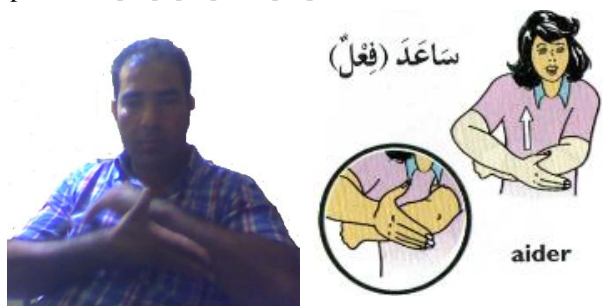

**Fig. 1.** A native Algerian signer (LSA sign 'Help').

In Sign Language there is a clear correspondence between meaning components of the event and the linguistic means (the hands and the body of the signer, and the space around him) employed to convey the event by the signer. The relation between form (linguistic means) and meaning is more explicit in Sign Languages than in Spoken languages [28]. Algerian deaf people have created their own Sign Language as in most communities around the world. However, as a result of long period of French occupation of Algeria, Algerian Sign Language (LSA) was influenced by French Sign Language (LSF). Actually, from time to time Algerian Sign Language teachers take some training sessions in French Sign Language; therefore, they have spread French Sign Language which has become mixed by Algerian Sign Language. In addition, the Council of Arab Ministers of Social Affairs (CAMSA), a committee within the League of Arab States (LAS) has approved the Unified Sign Language Dictionary [29]. A standard pan-Arab Sign Language (ArSL) drowns from different Arab Sign Languages.

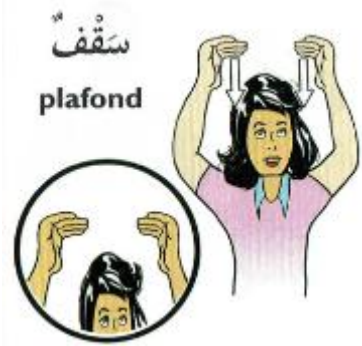

**Fig. 2.** The HandShape position change between the beginning location and the end location of the LSA Sign « سقف - ceiling ».

In order to promote Arabic Sign Language (ArSL) within the Arabic deaf community, Aljazeera channel uses ArSL in the signing TV journal. Whereas, some Sign language linguists are opposed to such attempts to unify Sign Languages in the Arab world, predicting thus the failure of the project. Nowadays, the Ministry of National Solidarity have launched a serious work to collect native Sign Language used by native Algerian deaf from different region of the country in order to constitute the Algerian Sign Language Dictionary, a support to be used in instruction of deaf students in educational systems. The last year (2017), the first edition of the dictionary was published (1500 Signs) (fig. 1, 2, 3, and 4).

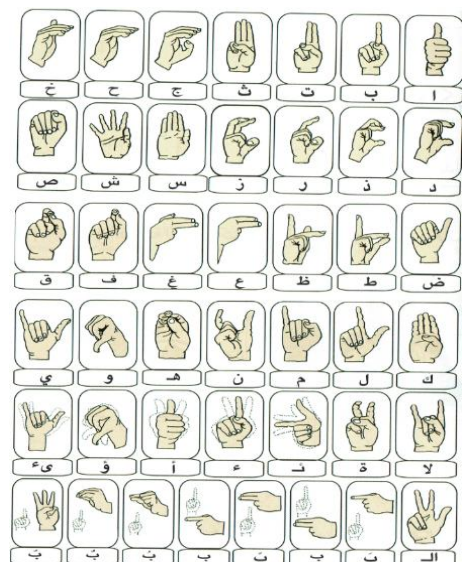

**Fig. 3.** The Algerian Sign Language (LSA) fingerspelled alphabet

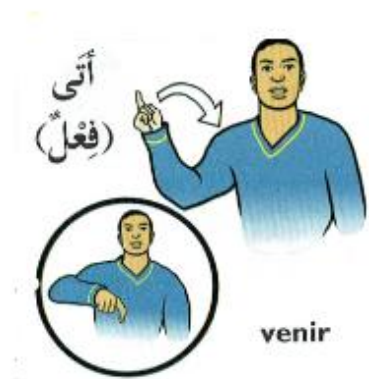

**Fig. 4.** The LSA Sign « أتى - come ».

Linguistically, in natural Sign languages, there are five parameters (chereme) employed to convey a message: (1) handshape(s); (2) location of sign; (3) palm orientation; (4) movement(s); and (5) non manual features (e.g., facial expressions, use of shoulders and body, and so on). Also, verb agreement in Sign Languages involves the linguistic use of space, it takes the following form: the beginning and ending points of the agreeing verb are associated with the points in space established for the arguments of the verb (subject, object). The signer establishes locations in the space around themselves ('signing space') to represent important referents in the discourse.

Sign language notations and transcription systems such as Stokoe notation [37] and Hamburg Notation System (HamNoSys) [38] use a set of written symbols to represent linguistic aspects of Sign Language in iconic way, whereas signs are instantiated in 3D space and need for spatial description. So, for modeling realistic signs, we have developed an X3D/VRML based visual tool with which we can represent morphological, syntactic and semantic features related to the designed sign, so that we can use it to automatically change some features if necessary. Information about sign features are transcribed in XML format from which, we can re-use existing simple signs in order to generate new complex sequences in systematic ways.

## IV. THE ARCHITECTURE OF THE SYSTEM

As our aim is to build a flexible Sign language generator that can produces avatar based Sign language contents from semantic Interlingua representation, the system requires some technical characteristics such as adaptability, portability and modularity. Our Sign language generator rests on state-of-the-art industry standards and resources, using X3D based multi-level notation for Signs transcription and H-Anim standard avatar anatomy and joints rotations. One reason behind the uses of X3D and H-Anim based signers is that most of free available avatars can be converted in these standard representations, giving an important means to adapt the system with regard deaf children preferences.

In this project, we aim to start with the widely used children's books in primary level education. The project can be subdivided into the following sub-goals:

1. Analyzing of Arabic sentences into semantic representation. Arabic sentences were extracted from school books and stored into xml based files.
2. Developing VR-based visual tool in order to synthesis signs in XML shared format using standard interchangeable 3D avatars.
3. Modeling and synthesizing sign according to the linguistic features unique to Sign Languages.
4. Generating Sign language from an Interlingua representation.

In this section, we focus mostly on the sign representation and generation system being proposed to provide rich description, semantically motivated system for the re-use and adaptability of sign entities. In order to generate 3d like human Sign sequences, our approach to sign representation [13] includes three levels of information: phonological level, syntactical level and semantic level. Phonological information includes a selected handshape (Fig. 3) from a standard set of handshapes instantiated in an initial space location and a movement toward a final location combined with non manuals signs (eye gaze, head tilt, and other body expressions). Also, the representation takes into account simultaneous and sequential aspects of signs. The choice of virtual environment is focused on the standard X3D (VRML 3rd generation) and H-Anim standard 2.0, enabling reuse of the features offered by its standards and better integration with the latest Web technologies. This enables the architecture to be embedded in a range of systems and platforms: the Avatar is readily displayed in most web-browsers and mobile computing devices.

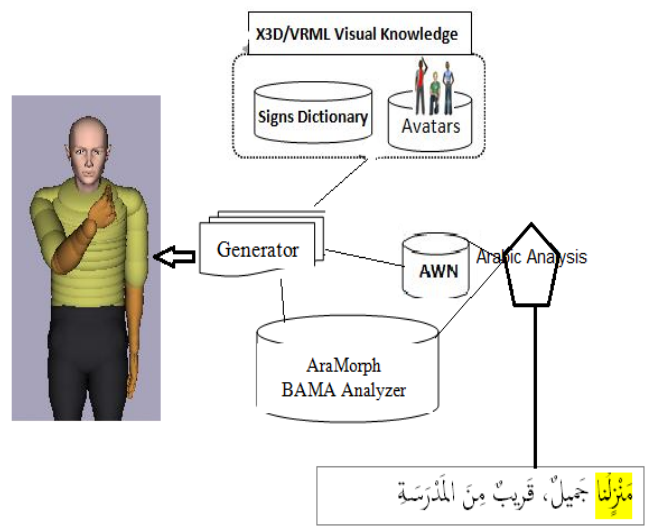

**Fig. 5.** An overview of the system architecture

This also means that the systems can be adapted to other languages, in particular English. Gestural entities generated are performed by a 3D computer character (avatar) according to the linguistic features unique to sign languages. The semantic content of a source Arabic sentence selected from the book text (Fig. 6.) is parsed and mapped into an internal semantic representation based on an Arabic version of *Framenet[14]*, to ensure accurate transfer of meaning of the sentence based on conceptual models from semantic-cognitive research. The Arabic *Framenet* semantic representation then drives the generation of gestural sequences by the virtual signer Avatar, using a dictionary of *FrameNet-to-X3D* mappings. We have, also, integrated the lexical semantic resource 'Arabic WordNet'[39] with its semantic network of Arabic words (verbs, nouns, adjectives etc.), in order to ensure a large coverage of signing words.

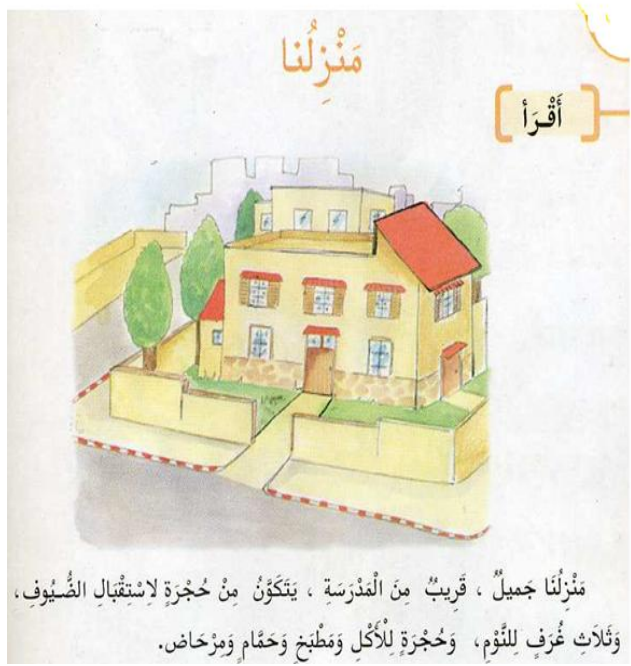

**Fig. 6.** A screenshoot of a page from school book.

The choice of virtual environment is focused on the standard X3D (VRML 3rd generation) and H-Anim standard 2.0, enabling reuse of the features offered by its standards and better integration with the latest Web technologies.

However, X3D/VRML rotations are designed in axis-angle representation. So, in order to realize a 3d Sign, we need to compute successive rotations and transitions of a set of joints that define the sign. Such a representation of rotations makes the signs conception task very hard and time consuming. Thus, we have integrated an animation module to our visual tool that allows several easy rotations methods such as successive yaw-pitch-roll rotation used by aviation systems. Also, we have integrated into our animation system the quaternion interpolation method to combine rotations. Thus, before the animation rendering starts, joints rotations are reordered by time and converted to quaternion in ordered to generate a smooth interpolation. Complex signs built from several basic signs undergo some internal changes with regard the time reference and the initial position of some joints.

The system (Fig. 7) provides a visual, flexible and interactive communication interface, which can be used as an educational tool for persons with hearing disabilities. We elaborate on the use of an Interlingua approach to produce the target signing sentence according to the linguistic features unique to Sign Language.

The design and implementation of our system follows a modular approach, involving two stages: (1) capturing Arabic text meaning into an Interlingua semantic representation, following a frame–based analysis [31] and taking into account linguistics information proper to Arabic; (2) Generating a signing contents corresponding to the intermediate semantic information in 3d environment using X3D and H-Anim parameters that actually animate the 3D Avatar signers to produce human like signs. The computer generated signing sentences were evaluated by native Algerian Sign Language signers.

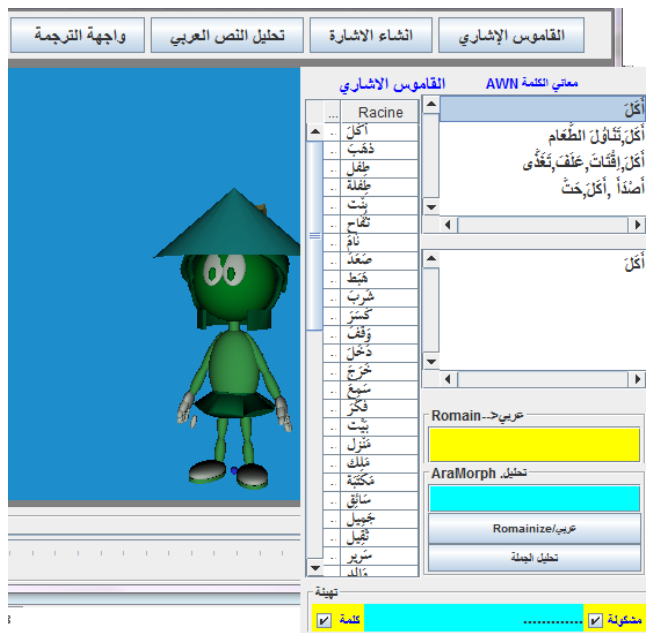

**Fig. 7.** A screenshot of the visual interactive environnement.

## V. CONCLUSION AND FUTUR WORKS

The most important contributions of this paper are: (1): The uses of Arabic WordNet words as part of our Arabic to Sign language Dictionary ; (2) the system uses X3D and H-Anim standards to generate 3D Sign language contents, thus they can be viewed via web navigators with VRML plug-in or HTML5 based technologies.

This work is a first stage; that aims to implement a gestural communication tool for the Algerian deaf pupils to facilitate their access to textual information written in standard Arabic. In the evaluation and looking forward work-package, we will plan what is required to scale up to coverage of the full primary level books, as a close approximation to full coverage of Arabic and LSA.

## REFERENCES

[1] É. Sáfár, I. Marshall, "The architecture of an English-text-to-Sign-Languages translation system", In *Recent Advances in Natural Language Processing (RANLP),* pp. 223-228, Tzigov Chark Bulgaria, 2001.204